\documentclass[journal]{IEEEtran}

\usepackage{graphicx}
\usepackage{algorithm}
\usepackage{array}
\usepackage{amsmath}
\usepackage{url}
\usepackage{enumerate}
\usepackage{multicol}
\usepackage{textcomp}
\usepackage{multirow}
\usepackage{cite}
\usepackage{fmtcount}
\usepackage{diagbox}

\usepackage{color}
\definecolor{Change}{rgb}{0.5,0.1,0.1}

\usepackage{threeparttable}

\usepackage{changes}
\definechangesauthor[name={Per cusse}, color=orange]{per}
\setremarkmarkup{(#2)}
\usepackage[noend]{algpseudocode}
\usepackage{algorithm}
\usepackage{algorithmicx}
\usepackage{varwidth}

\begin{document}
	%\title{Semantic Segmentation of 3D point clouds using Semi-supervised Learning}
	\title{Semantic Segmentation of 3D LiDAR Data in Dynamic Scene Using Semi-supervised Learning}
	
	\author{Jilin~Mei,~\IEEEmembership{Member,~IEEE,}
		Biao~Gao,~\IEEEmembership{Member,~IEEE,}
		Donghao~Xu,~\IEEEmembership{Member,~IEEE,}
		Wen~Yao,~\IEEEmembership{Member,~IEEE,}
		Xijun~Zhao,~\IEEEmembership{Member,~IEEE,}
		and Huijing~Zhao, ~\IEEEmembership{Member,~IEEE,}
		
		\thanks{This work is supported by the National Key Research and Development Program of China (2017YFB1002601) and the NSFC Grants (61573027).}
		\thanks{J. Mei, B. Gao, D. Xu and H. Zhao are with the Peking University, with
			the Key Laboratory of Machine Perception (MOE), and also with the School of
			Electronics Engineering and Computer Science}% <-this % stops a space
		\thanks{W. Yao and X. Zhao is with China North Vehicle Research Institute, Beijing, China.}% <-this % stops a space
		\thanks{Correspondence: H. Zhao, zhaohj@cis.pku.edu.cn.}
		}

	%\markboth{IEEE TRANSACTIONS ON INTELLIGENT TRANSPORTATION SYSTEMS,~Vol.~?, No.~?, AUGUST~2018}%
	%{Shell \MakeLowercase{\textit{et al.}}: IEEE TRANSACTIONS ON INTELLIGENT TRANSPORTATION SYSTEMS}

	\maketitle
	
	\begin{abstract}
	This work studies the semantic segmentation of 3D LiDAR data in dynamic scenes for autonomous driving applications. A system of semantic segmentation using 3D LiDAR data, including range image segmentation, sample generation, inter-frame data association, track-level annotation and semi-supervised learning, is developed. To reduce the considerable requirement of fine annotations, a CNN-based classifier is trained by considering both supervised samples with manually labeled object classes and pairwise constraints, where a data sample is composed of a segment as the foreground and neighborhood points as the background. A special loss function is designed to account for both annotations and constraints, where the constraint data are encouraged to be assigned to the same semantic class. A dataset containing 1838 frames of LiDAR data, 39934 pairwise constraints and 57927 human annotations is developed. The performance of the method is examined extensively. Qualitative and quantitative experiments show that the combination of a few annotations and large amount of constraint data significantly enhances the effectiveness and scene adaptability, resulting in greater than 10\% improvement.
	\end{abstract}

	
	\section{Introduction}
\begin{figure}
		\centering
		\includegraphics[width=0.5\textwidth]{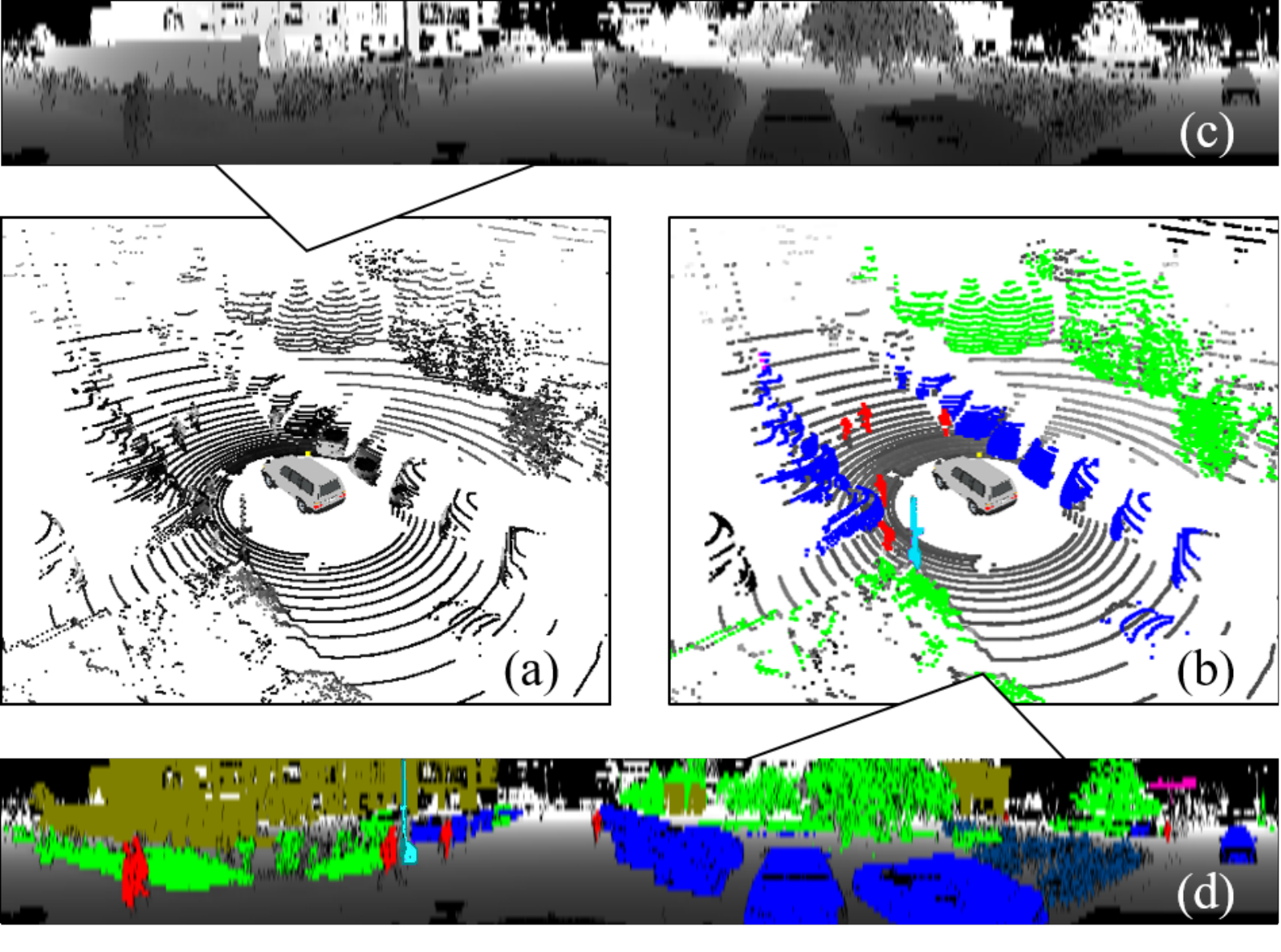}
		\caption{The semantic segmentation for dynamic scene. (a) and (c) show the input data in two kinds of formats, i.e., the raw 3D point clouds and range frame. (b) and (d) show the semantic segmentation results.}
		\label{fig:dynamiccampus}
	\end{figure}

	Scene understanding is crucial for the safe and efficient navigation of autonomous vehicles in complex and dynamic environments, and semantic segmentation is a key technique. 3D LiDAR has been used as one of the main sensors in many prototyping systems for fully autonomous driving\cite{urmson2008autonomous}. Semantic segmentation using 3D LiDAR data is illustrated in Fig. \ref{fig:dynamiccampus}, where given a frame of input data (a), the problem is to find a meaningful label (i.e., object class in this research) for each pixel, super-pixel or region of the data (b). As 3D LiDAR is a 2.5D sensing of the surroundings, it can be represented equivalently in the form of a range image (c)-(d) in the polar coordinate system, and the problem of semantic segmentation can be solved by using either 3D points or range images as the input.
	
	Semantic segmentation using 3D LiDAR data from outdoor scenes has been studied since the past decade \cite{urmson2008autonomous,moosmann2009segmentation,douillard2011segmentation}. The traditional process in these works \cite{munoz2009onboard,zhao2010scene} includes the following steps: (1) preprocessing to divide a whole dataset into locally consistent small units, such as voxels, segments or clusters; (2) extracting a sequence of predefined features; (3) learning a classifier via SVM, random forest etc.; and (4) refining the results using a method such as conditional random field by considering the spatial consistency among neighboring units. The traditional methods depend on carefully designed discriminative features, and their adaptability to different scenes remains an open challenge. 
	
	The recent success of deep learning in image semantic segmentation has provided new approaches\cite{garcia2017review}. These methods remove the dependence on handcrafted features in an end-to-end manner. However, these methods also have substantial demands for finely labeled data\cite{garcia2017review}. On one hand, pixel-wise annotation is extremely time consuming; on the other hand, few 3D LiDAR datasets with annotation at the point level to support autonomous driving applications are publicly available. Therefore, it is necessary to develop a semantic segmentation method using 3D LiDAR data with only a small set of supervised data, where semi-supervised learning is adopted.
	
	Semi-supervised learning methods, which integrate labeled and unlabeled data, have been studied extensively in the field of machine learning \cite{zhu2006semi}. In \cite{lange2005learning,lu2005semi}, pairwise constraints are collected from unlabeled data to describe the probability of two samples sharing the same or different labels. LiDAR sensors measure the 3D coordinates of an object directly and can be used to associate the same object in the data of subsequent frames according to their locations after ego motion compensation. A tracking method can be used for data association for moving objects such as people, cyclists and cars. Such associated data constitute pairwise constraints, which can be inexpensive due to their abundant nature and autonomous generation.
	
	\begin{table*}[]
		\centering
		\renewcommand\arraystretch{1.5}
		%\small
		\begin{threeparttable}
			\caption{Approaches for 3D point clouds semantic segmentation}
			\label{introtabel}
			\begin{tabular}{|cccccc|l}
				\cline{1-6}
				Research                                     & Learning Method & Input & Classifier              & Dataset                                                                     & Scene                           &  \\ \cline{1-6}
				Munoz, 2009,\cite{munoz2009onboard}           & sup.            & L     & MRF                     & -                                                                           & rural                           &  \\
				Zhao, 2010,\cite{zhao2010scene}               & sup.            & L     & SVM                     & -                                                                           & campus                          &  \\				
				Weinmann, 2014\cite{weinmann2014semantic}     & sup.            & L     & RF                      & VMR-Oakland\cite{munoz2009contextual},Pairs-rue-Madame\cite{serna2014paris} & urban                           &  \\
				Munoz, 2009,\cite{munoz2009contextual}        & sup.            & L/V   & MRF                     & -                                                                           & -                               &  \\															
				Hackel, 2016\cite{hackel2016fast}             & sup.            & L     & RF                      & Pairs-rue-Madame                                                            & urban                           &  \\
				Hu, 2013,\cite{hu2013efficient}               & sup.            & L     & KLR                     & VMR-Oakland,Freiburg\cite{behley2012performance}                            & urban                           &  \\
				Lu, 2012,\cite{lu2012simplified}              & sup.            & L     & CRF,SVM                 & VMR-Oakland                                                                 & urban                           &  \\ \cline{1-6}
				Engelmann, 2017,\cite{engelmann2017exploring} & sup.            & L/V   & DL                      & S3DIS\cite{armeni_cvpr16},vKITTI\cite{GaidonCVPR2016},KITTI                 & indoor,urban,urban              &  \\
				Tosteberg, 2017,\cite{tosteberg2017semantic}  & sup.            & L     & DL                      & Semantic3D,VPS\cite{vpsdataset}                       & urban,indoor                    &  \\
				Dewan, 2017,\cite{dewan17iros}                & sup.            & L     & DL                      & KITTI\cite{Geiger2013IJRR}                                                  & urban                           &  \\
				Hackel, 2017,\cite{hackel2017isprs}           & sup.            & L     & DL                      & Semantic3D\cite{hackel2017isprs}                                                                  & urban/rural                     &  \\
				Wu, 2017,\cite{wu2017squeezeseg}              & sup.            & L     & DL                      & KITTI						                                                & urban                           &  \\
				Piewak, 2018,\cite{piewak2018boosting}        & sup.            & L/V   & DL                      & -                                                                           & urban/rural/highway             &  \\
				Caltagirone, 2017,\cite{caltagirone2017fast}  & sup.            & L     & DL                      & KITTI  						                                                & urban                           &  \\
				Lawin, 2017,\cite{lawin2017deep}              & sup.            & L     & DL                      & Semantic3D                                                                  & urban/rural                     &  \\
				Tchapmi, 2017,\cite{tchapmi2017segcloud}           & sup.            & L/V   & DL                      & NYU V2\cite{silberman2012indoor},Semantic3D,S3DIS,KITTI,                    & indoor,urban/rural,indoor,urban &  \\
				Riegler, 2017,\cite{riegler2017octnet}        & sup.            & L     & DL                      & ModelNet10\cite{wu20153d}                                                   & CAD model                       &  \\
				Qi, 2017,\cite{qi2017pointnet}                & sup.            & L/V   & DL                      & ShapeNet\cite{yi2016scalable},Stanford3D\cite{armeni20163d}                 & CAD model,indoor                &  \\
				Landrieu, 2017,\cite{landrieu2017large}       & sup.            & L/V   & DL                      & Semantic3D,S3DIS                                                            & urban,rural/indoor              &  \\ \cline{1-6}
				Bearman, 2016,\cite{bearman2016s}             & semi-sup.       & V     & DL                      & PASCAL VOC\cite{everingham2010pascal}                                                                  & -                               &  \\
				Yan, 2006,\cite{yan2006discriminative}        & semi-sup.       & V     & KLR,SVM                 & -                                                                           & nursing home                    &  \\
				Bauml, 2013\cite{bauml2013semi}               & semi-sup.       & V     & KLR,SVM                 & -                                                                           & TV series                       &  \\
				Hong, 2015,\cite{hong2015decoupled}           & semi-sup.       & V     & DL                      & PASCAL VOC                                     & -                               &  \\
				Papandreou, 2015,\cite{papandreou2015weakly}  & semi-sup.       & V     & DL                      & PASCAL VOC                                                                  & -                               &  \\
				Lin, 2016,\cite{lin2016scribblesup}           & semi-sup.       & V     & DL                      & PASCAL VOC                                                                  & -                               &  \\
				Cour, 2009,\cite{cour2009learning}            & weakly-sup.      & V     & linear classifier & -                                                                           & TV series                       &  \\
				Pathak, 2015,\cite{pathak2015constrained}     & weakly-sup.      & V     & DL                      & PASCAL VOC                                                                  & -                               &  \\
				Xu, 2015,\cite{xu2015learning}                & weakly-sup.      & V     & linear classifier  & Siftflow\cite{liu2011nonparametric}                                         & -                               &  \\
				Dai, 2015,\cite{dai2015boxsup}                & weakly-sup.      & V     & DL                      & PASCAL VOC                                                                  & -                               &  \\ \cline{1-6}
			\end{tabular}
			\begin{tablenotes}
				\footnotesize
				\item  sup. : supervised; L : LiDAR Data; V : Camera Data;  RF : Random Forest; KLR : Kernel Linear Regression; DL : Deep Learning.
			\end{tablenotes}
		\end{threeparttable}
	\end{table*}	
	
	In this paper, we propose a semantic segmentation using 3D LiDAR data from dynamic urban scenes by integrating semi-supervised learning and deep learning methods. A CNN-based classifier is trained by considering both supervised samples with manually labeled object classes and pairwise constraints, where a data sample is composed of a segment as the foreground and the neighborhood points as the background. A system of semantic segmentation using 3D LiDAR data, including range image segmentation, sample generation, inter-frame data association, track-level annotation and semi-supervised learning, is developed. A dataset containing 1838 frames of LiDAR data, 39934 pairwise constraints and 57927 human annotations is generated using 3D LiDAR data from a dynamic campus scene. Qualitative and quantitative experiments show that the combination of a small amount of annotation data and a large amount of constraint data significantly improves the effectiveness and scene adaptability of the classifier.
	
	%This work provides the following contributions: (1) A new semi-supervised semantic segmentation method are proposed for LiDAR data collected from driving platform. (2) The pairwise constraint is introduced in the training phase to reduce the large requirement of labeled data. (3) the performance of our method is demonstrated on a new dataset, which will be provided together with this publication.
	
	The remainder of this paper is organized as follows. Related work is discussed in Sect. \uppercase\expandafter{\romannumeral2}. In Sect. \uppercase\expandafter{\romannumeral3}, the proposed method is presented. In Sect. \uppercase\expandafter{\romannumeral4}, the algorithm details are discussed. Then, we present the experimental results in Sect. \uppercase\expandafter{\romannumeral5} and draw conclusions in Sect. \uppercase\expandafter{\romannumeral6}.

	\section{Related Works}
	Numerous studies on semantic segmentation have been conducted; the recent progress for image and RGB-D data is reviewed in \cite{garcia2017review}. In this section, we focus on methods specified for 3D point clouds (i.e., from LiDAR sensors) in dynamic outdoor scenes. These works can be broadly divided into three classes: feature-based methods, deep learning methods, semi-supervised learning methods.

	\subsection{Feature-based Methods}
	Feature-based methods belong to traditional machine learning, and the general process of these methods consists of feature selection, classifier design and graphical model description.
	
	A straightforward technique is to convert semantic segmentation into a point-wise classification that includes extracting the features on each unit, concatenating the features as a vector and determining the label via a well-trained classifier. \cite{weinmann2014semantic} presents a common pipeline from feature selection to classifier training. Due to the irregular arrangement of point clouds, the authors test 5 definitions of neighborhood to achieve the best representation. Similar research is conducted in \cite{hackel2016fast}, which demonstrates the ability to address varying densities of data. A single point cloud usually contains millions of points, so evaluating the label for each point is typically computationally expensive (on the order of minutes according to \cite{hackel2016fast}). 
	
	\cite{hu2013efficient} proposes an efficient approach where speed and accuracy are satisfied simultaneously; furthermore, the average classification time can be reduced to less than 1 s. \cite{zhao2010scene} represents the raw point cloud as a 2D range image and proposes a framework for simultaneous segmentation and classification of the range image that considers both the 2D range image and 3D raw data. Straightforward approaches assume that each data unit is independent and ignore the spatial and contextual relations between units. Consequently, they can produce good results based on distinctive features. However, when the features are not discriminative, the point-wise classification will be noisy and locally inconsistent\cite{munoz2009onboard}.
	
	The neighbor elements are taken into account to make the segmentation results spatially smooth. For this purpose, graphical models such as Markov random Field (MRF) and conditional random field (CRF) are usually exploited to encode the spatial relationships. In \cite{munoz2009contextual}, the node potentials and edge potentials are both formulated with a parametric linear model, and the functional max-margin learning is used to find the optimal weights. \cite{lu2012simplified} proposes a simplified Markov network to infer the contextual relations between points. Instead of learning all the weights for the node and edge potentials in graphical models, the node potentials are calculated from a point-wise classifier, and the edge potentials are determined by the physical distance between points. 
	
	%In the following works, researchers mainly focus on how to define a node and how to build the graph. In \cite{guinard2017weakly}, both local and non-local descriptors are designed, furthermore, the scene is partitioned into geometrically-homogeneous segments, then the adjacency graph is build and a CRF classifier is designed to capture the high-order spatial relations. The previous graphical models here only encourage neighboring nodes to have the same label, however, \cite{najafi2014non} introduce a new higher-order model for 3D point cloud classification that takes into account the non-associative geometric context between different labels. Graphical model is not the only way to incorporating contextual information. \cite{xiong20113} adopts a sequenced/hierarchical prediction method at different scales so that the contextual information can be transfered between points and regions.
	
	The performance of the above methods largely depends on handcrafted features. These methods are effective in fixed or regular scenarios, but for dynamic scenes, the features are empirically designed and the performance decreases.

	\subsection{Deep Learning Methods}
	Deep learning, especially the convolution neural network (CNN) without handcrafted features, has shown effectual performance on 2D image segmentation\cite{garcia2017review}. However, the semantic segmentation of 3D point clouds(i.e., from LiDAR sensors) is still an open research problem\cite{engelmann2017exploring} due to the irregular, not grid-aligned properties. Therefore, recent studies project the point clouds into 2D views, and some of them attempt to directly ways, for example, volumetric/voxel representations.
		
	Inspired by the success of CNN in image segmentation, the state-of-the-art image-based algorithms can be used directly after rendering 2D views from the 3D raw data. \cite{tosteberg2017semantic} projects point clouds into virtual 2D RGB images via Katz projection. Then, a pretrained CNN is used to semantically classify the images. However, this projection removes all the points that are not visible; for example, if a car is projected, all the points behind it are removed. \cite{dewan17iros} unwraps $360^\circ$ 3D LiDAR data onto a spherical 2D plane without point loss. Spherical projection is also applied in SqueezeSeg\cite{wu2017squeezeseg}, where the CNN directly outputs the point-wise label of the transformed LiDAR data and a CRF is applied to refine the outputs. \cite{piewak2018boosting} uses cylindrical projection to create the depth and reflectivity images. In \cite{caltagirone2017fast}, the point clouds are encoded by top-view images and a simple fully convolutional neural network (FCN) is used. This method can be used in real time because only elevation and density features are extracted. In \cite{lawin2017deep}, the input point cloud is projected into multiple views, such as color, depth and surface normal images. 
	
	Another type of method models the raw data in direct ways. \cite{tchapmi2017segcloud} proposes SEGCloud, where the raw 3D point cloud is preprocessed into a voxelized point cloud with a fixed grid size. Although \cite{tchapmi2017segcloud} is simple and effective, how to set the voxel size is a problem in large-scale scenes. Thus, the scene is voxelized at five different resolutions in \cite{hackel2017isprs}, and each of the five scales is handled separately by the CNN. Rather than using a fixed grid size, \cite{riegler2017octnet} proposes OctNet, where the hybrid grid-octree data structure is applied to represent the raw 3D data, and each leaf of the octree stores a pooled feature representation. PointNet\cite{qi2017pointnet} is a unified architecture that directly takes raw point clouds as input and outputs the label of each point. The scene is divided into blocks. Then, the points in each block are passed through a series of multilayer perceptrons (MLPs) to extract the local and global features. Based on \cite{qi2017pointnet}, \cite{engelmann2017exploring} extends the method to incorporate a larger-scale spatial context, and improved results are reported in both indoor and outdoor scenarios. \cite{landrieu2017large} proposes a more elegant architecture to capture contextual relations. The first step is to partition the raw point cloud into geometrically simple shapes, called super-points. The super-points are then embedded by PointNet\cite{qi2017pointnet}.

	Semantic segmentation with deep learning is usually implemented in a supervised manner, which requires detailed annotations. However, obtaining point-wise annotations for 3D point clouds is labor intensive and time consuming. Furthermore, few public datasets support this level of annotation.  

		\begin{figure*}
			\centering
			\includegraphics[width=0.8\textwidth]{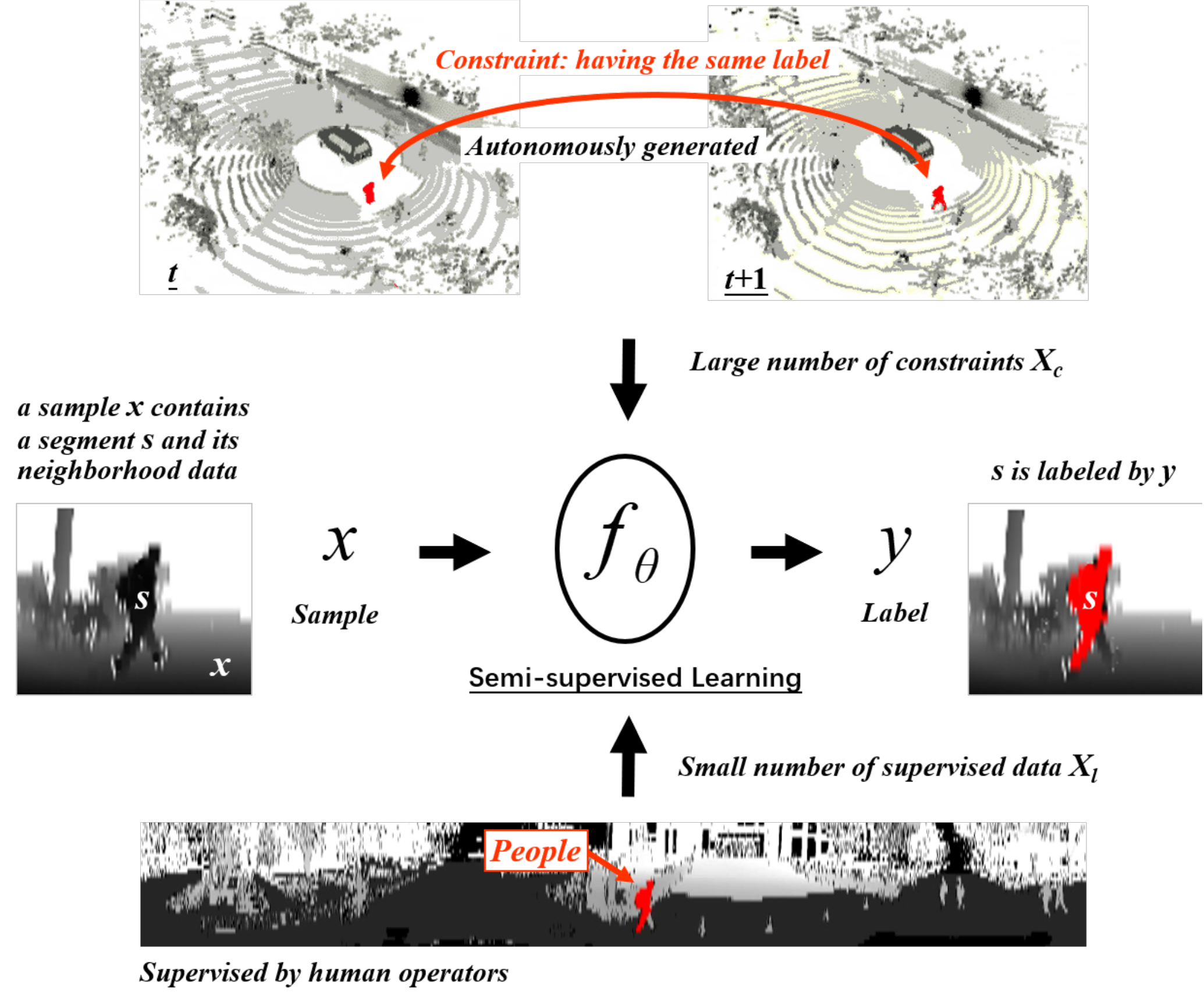}
			\caption{The framework of semi-supervised learning for 3D point clouds  semantic segmentation.}
			\label{fig:algframework}
		\end{figure*}

	\subsection{Semi-supervised Learning Methods}
	Considering the large demand for detailed annotations, many researchers study semi-supervised learning methods, which integrate fewer labeled data and more unlabeled data, and weakly supervised learning methods, which use multiple ambiguous labels. Our research belongs to the semi-supervised category, please refer to TABLE  \ref{introtabel} for weakly supervised methods.
	
	The early work \cite{lange2005learning} on semi-supervised learning specifies the prior knowledge of unlabeled data via pairwise constraints. A pairwise must-link constraint means two objects must have the same label, although the label is unknown, whereas two objects associated via must-not-link must have different labels. Both labeled and constraint data are used for model fitting, and the authors model the constraint information by the maximum entropy principle. A similar idea is presented in \cite{lu2005semi}. The pairwise constraints are incorporated in the clustering of a Gaussian mixture model (GMM). \cite{lange2005learning,lu2005semi} present the foundation of semi-supervised learning with pairwise constraints.	

	\cite{yan2006discriminative} extends the method for video object classification, where temporal relations between frames, multi-modalities, such as faces and voices, and human feedback (manual annotation) are considered and formulated in a unified framework. \cite{bauml2013semi} applies constraints for person identification in multimedia data and achieves state-of-the-art performance on two diverse TV series. Recently, semi-supervised learning has also been integrated with deep neural networks (DNN). \cite{hong2015decoupled} decouples semantic segmentation into classification and segmentation using a large number of image-level object labels and a small number of pixel-wise annotations. The classification network specifies the class-specific activation maps, which are transfered into the segmentation network with bridge layers. Then, the segmentation network requires only a few pixel-wise annotations to train the model, e.g., 5 or 10 strong annotations per class. \cite{papandreou2015weakly} designs a expectation-maximization (EM) training method for semantic image segmentation by combining bounding boxes, image-level labels and a few strongly labeled images.
	
	Semi-supervised learning has been successfully applied in image segmentation\cite{hong2015decoupled} and video analysis\cite{yan2006discriminative}. However, to the best of the author's knowledge, few studies discuss semi-supervised leaning in the context of the 3D point clouds in semantic segmentation. In this paper, we attempt to combine semi-supervised learning and neural network to solve the problem of how to perform 3D point cloud semantic segmentation with insufficient point-wise annotations. 

	\begin{figure*}
		\centering
		\includegraphics[width=0.8\textwidth]{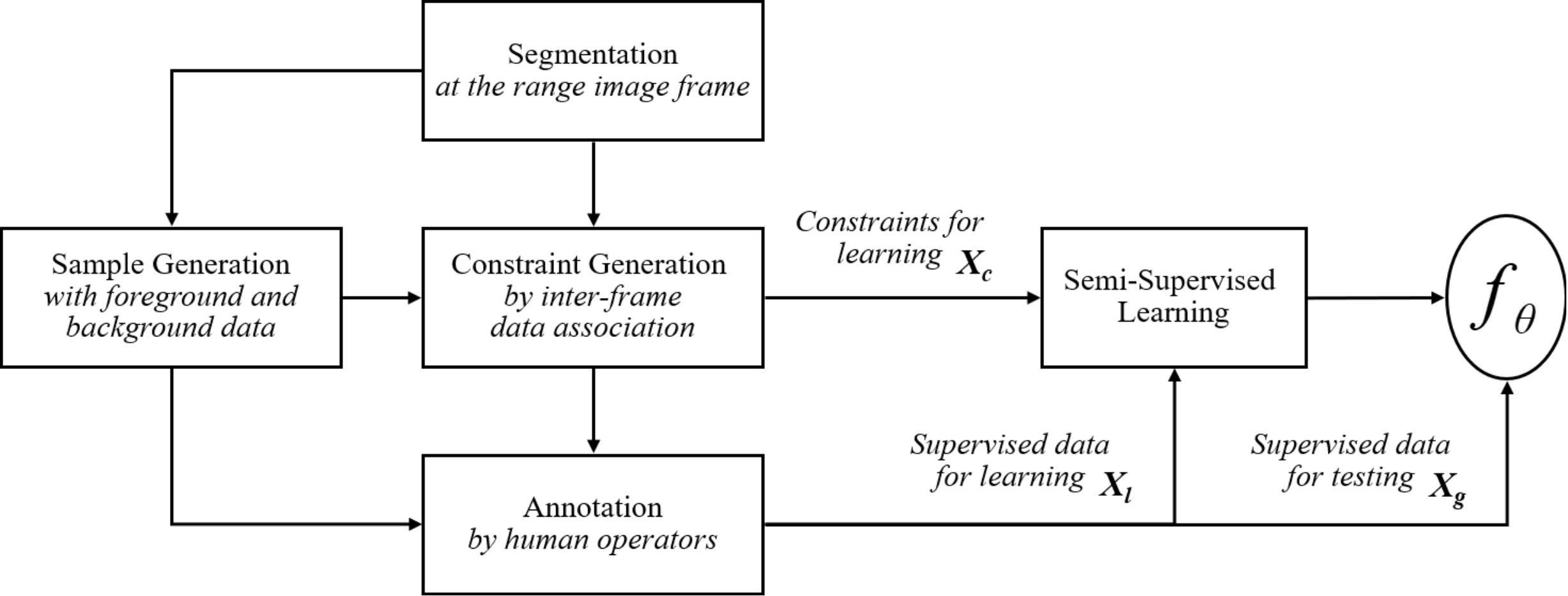}
		\caption{The flowchart of implementation.}
		\label{fig:implement_flowchart}
	\end{figure*}
	
\section{METHODOLOGY}
	\subsection{Problem Definition}
	
	Let $s$ denotes a small segment of 3D LiDAR data extracted by examining the consistency of 3D points with their neighborhood in the range image frame using, e.g., a region growing method. Without loss of generality, we assume that the 3D points of $s$ are measurements of a single object. 
	However, $s$ commonly represents only a part of the object, e.g., the upper body of a pedestrian, due to the nature of LiDAR measurements.
	Hence, for each $s$, a data sample $x$ is generated centered at $s$, containing $s$ as the foreground and its neighborhood data as the background, as shown in Fig. \ref{fig:algframework}.
	The problem in this work is formulated as learning a multi-class classifier $f_\theta$ that maps $x$ to a label $y \in \{1,...,K\}$ and subsequently associates $y$ with the 3D points of $s$.
	
	\begin{equation}
	f_{\theta}: x\to y \in \{1,...,K\}
	\end{equation}
	
	Given a set of supervised data samples $X_l=\{x_i, y_i\}_{i=1}^M$, where $\{y_i\}$ are one-hot vectors annotated manually by human operators for each $\{x_i\}$, a common way of learning a classifier $f_\theta$ is to find the best $\theta^*$ that minimizes a loss function $L$, as below.
	
	\begin{equation}
	\theta^{*}=\mathop{\arg\max}_{\theta}L(X_l; \theta)
	\end{equation}
	
	However, the problem of generating a large amount of supervised data is not trivial. This research learns $f_\theta$ with a small set of costly supervised data $X_l$ and an additional large set of autonomously generated constraints $X_c=\{<x_{i},x_{j}>_n\}^{N}_{n=1}$,
	
	\begin{equation}
	\theta^{*}=\mathop{\arg\max}_{\theta}L(X_l, X_c;\theta)
	\end{equation}
	where constraint $<x_{i},x_{j}>$ denotes that $x_{i}$ and $x_{j}$ have the same label, i.e., $y_i=y_j$, which is generated in this research autonomously by associating the data segments along sequential frames according to their locations after ego-motion compensation. 
	
	This semi-supervised loss function $L$ is based on a combination of loss on supervised data $X_l$ and loss on constraints $X_c$.
	
	\begin{equation}
	L(X_l, X_c; \theta) = L_l(X_l; \theta) + L_c(X_c; \theta)
	\end{equation}

	\begin{figure*}
		\centering
		\includegraphics[width=0.9\textwidth]{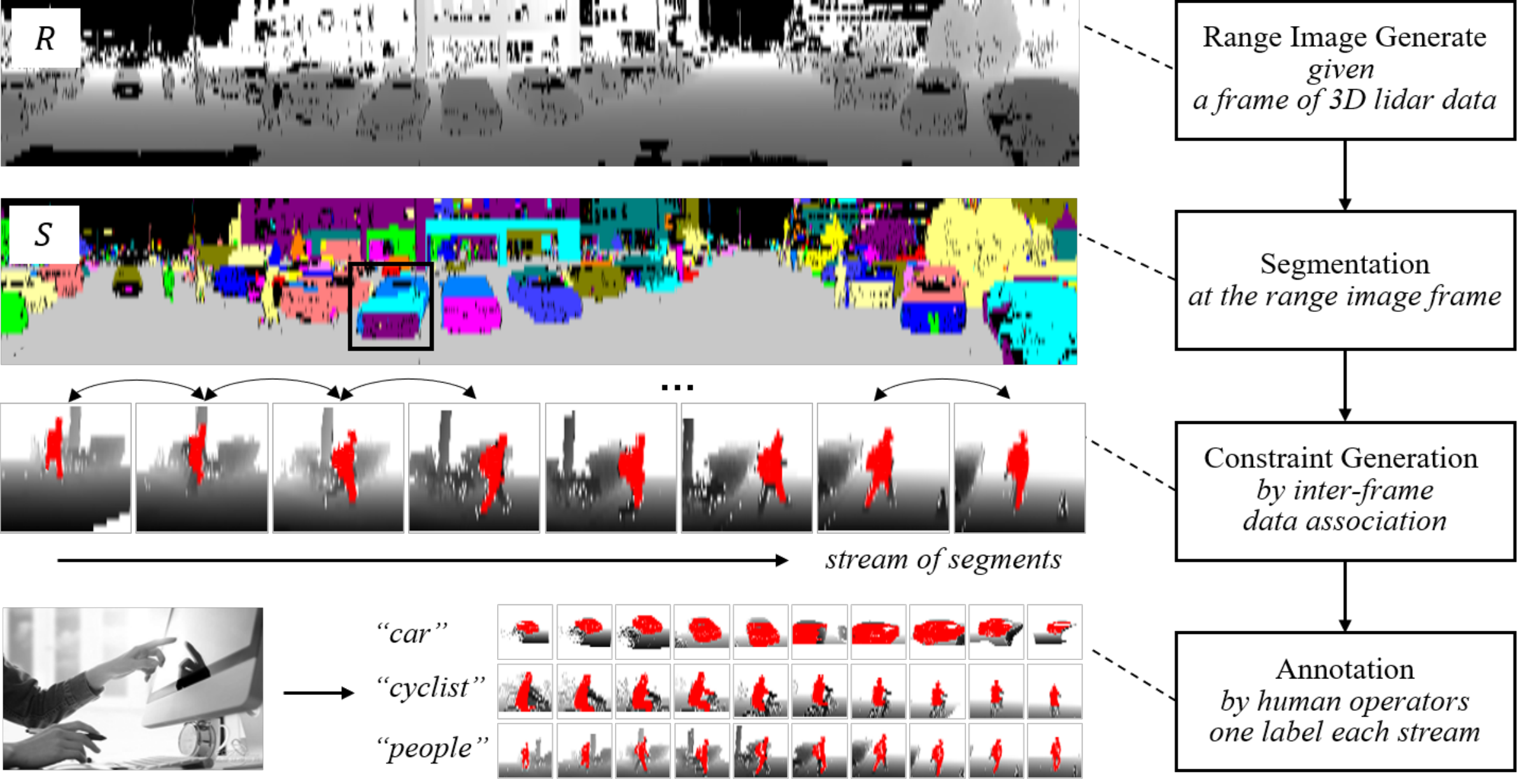}
		\caption{The details for segmentation, constraint generation and annotation. The first row is range image; the second row is the segmentation result and the black rectangle shows a car is separated into three parts; the third row shows two adjacent sample consist of one constraint; the fourth row is human annotation.}
		\label{fig:implement_details}
	\end{figure*}
	
	\subsection{Supervised Loss}
	For supervised data $X_l$, we follow the widely used definition of cross entropy, e.g., \cite{qi2017pointnet}, and define loss $L_l$ as below:
	\begin{equation}
	L_l(X_l;\theta)=-\frac{1}{M}\sum_{i=1}^M\sum_{k=1}^K{\textbf{1}[y^{k}_{i}=1]ln(P^k_{\theta}(x_i))}
	\end{equation}
	where $\textbf{1}[*]$ is an indicator function, and $P^k_{\theta}(x_i)$ is the probability that $x_i$ is assigned a label $k$ by a classifier with the set of parameters $\theta$.
	
	\subsection{Constraint Loss}
	
	For each constraint $<x_{i},x_{j}>$, let $y_i$ and $y_j$ denote the labels $x_{i}$ and $x_{j}$, respectively, by using a learned classifier $f_{\theta}$. A penalty is applied if $y_i \neq y_j$. Subsequently, the loss is estimated as below for each constraint based on the probability that the constrained data samples are assigned different labels.	
	\begin{equation}
	\begin{split}
	P(y_{i}\neq y_{j})&=\sum_{k=1}^K\sum_{\substack{l=1 \\ l\neq k}}^KP^k_{\theta}(x_i)P^l_{\theta}(x_j) \\
	&=1-\sum_{k=1}^KP^k_{\theta}(x_i)P^k_{\theta}(x_j)
	\end{split}
	\end{equation}
	
	Hence, we define the loss on constraints $L_c$ as below:
	\begin{equation}
	\begin{split}
	L_{c}(X_c;\theta)&=\frac{\gamma}{N}\sum_{n=1}^NP(y_{i}\neq y_{j}) \\
	&=\frac{\gamma}{N}\sum_{n=1}^N(1-\sum_{k=1}^KP^k_{\theta}(x_{i})P^k_{\theta}(x_{j})), \gamma \in [0,1]
	\end{split}
	\end{equation}	
	where $\gamma$ is a weighting factor. Clearly, $L_c$ describes the unsupervised learning procedure.
	
	\subsection{Semi-supervised Loss}
	Combining the losses from supervised data $X_l$ and constraints $X_c$, the semi-supervised loss is defined as below.
	\begin{equation}
	\label{eq:loss}
	\begin{split}
	L(X_l,&X_c;\theta)=L_l(X_l;\theta)+L_c(X_c;\theta) \\
	&=-\frac{1}{M}\sum_{i=1}^M\sum_{k=1}^K{\textbf{1}[y^{k}_{i}=1]ln(P^k_{\theta}(x_i))} + \\
	&\frac{\gamma}{N}\sum_{n=1}^N(1-\sum_{k=1}^KP^k_{\theta}(x_{i})P^k_{\theta}(x_{j})), \gamma \in [0,1].
	\end{split}
	\end{equation}

\section{Algorithm Details}

	\subsection{Process Flow}
	Fig. \ref{fig:implement_flowchart} describes the major modules of the workflow. Sample generation and semi-supervised learning are detailed in the next subsections. Here, we describe the remaining modules. Although traditional methods are used in these modules, their integration and the design of the data pipelines are important in constructing a complete system. 
	
	As illustrated in Fig. \ref{fig:implement_details}, segmentation is conducted at the frame level of a range image, which is a representation of 3D LiDAR data in the polar coordinate system. The columns and rows of a range image correspond to tessellated horizontal and vertical angles, and each pixel value is the range distance of the laser point in that direction. A Velodyne HDL-32E is used in this research. Hence, a LiDAR frame contains 32 scan lines at different vertical angles, and each scan line has 2160 laser points at different horizontal angles. These laser points are projected onto a range frame and reshaped to size (144,1080).
	
	Region growing is conducted to extract segments from unlabeled pixels as the seeds by examining 4-connectivity, where the thresholds on the vertical and horizontal range differences of two connected pixels are assigned empirically with a set of test data. As detailed in the next subsection, a segment is used as the foreground data in sample generation, and segmentation is treated only as a preprocessing step. Many methods can be exploited as long as the following condition is met: the 3D points of segment $s$ are measurements of a single object. 
	
	Constraints are generated by inter-frame data association. For any segment $s_t$ in frame $t$, the 3D points can be back-projected to the LiDAR sensor's coordinate system in the previous frame $t-1$ based on the vehicle's motion data and calibration parameters. A segment $s’_{t-1}$ is associated with $s_t$ if it matches with the 3D points. Let $x_i$ and $x_j$ denote the samples of $s’_{t-1}$ and $s_t$, respectively; a pairwise-constraint $\langle x_i, x_j\rangle$ is subsequently generated.
	
	As illustrated in Fig. \ref{fig:implement_details}, data association is conducted for the entire stream, and sequences of segments are extracted as the result. An operator examines each sequence. If the sequence is tracked correctly, a label is assigned to all the segments (and the samples) of the sequence, and constraints are generated for all subsequent samples. Otherwise, a sequence is manually truncated to remove the erroneous tracking part or even discarded.
	
	The dataset with supervised labels is divided into two groups, $X_l$ and $X_g$, for training and testing, respectively. The set of pairwise constraints $X_c$ is used for training only.

	\begin{figure*}
		\centering
		\includegraphics[width=1.0\textwidth]{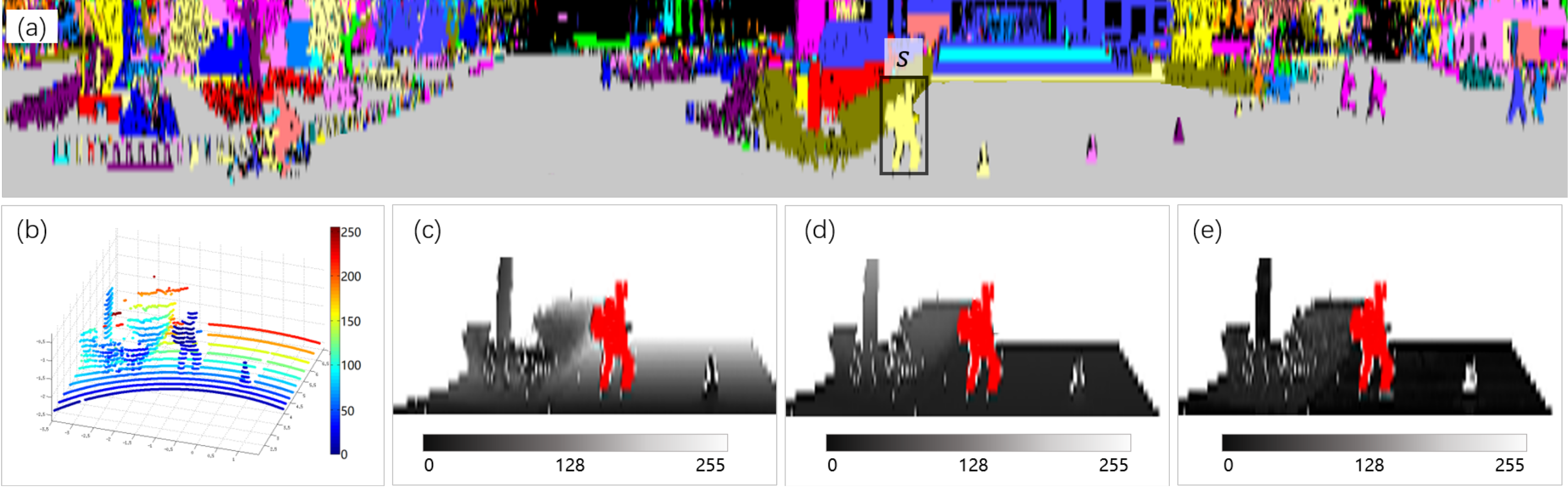}
		\caption{The procedure of sample generation. (a) the yellow segment $s$ in the black rectangle is chosen as candidate region. (b) the raw points inside a cuboid centered at $s$ are cropped, where the cuboid size is 2.4mx5mx5m(height,width,length) and the points are colored by range value; then these points are projected on range image to make one sample that consists of three channels. (c) the range channel of the sample, and we mark $s$ with red for better visualization. (d) the height channel of the sample. (e) the intensity channel of the sample.}
		\label{fig:implement_sample}
	\end{figure*}
	
	\subsection{Sample Generation}
	A straightforward method is to use a segment $s$ directly as the input of a classifier, i.e., $s=x$. However, the performance of such a classifier can be degraded if $s$ represents only a part of the target object. This situation often occurs in the segmentation results of 3D LiDAR data. Due to diffusive reflection or the weak reflectance of LiDAR measures on reflective or dark objects, many failures that yield discontinuities in the data of a single object exist in 3D LiDAR measurements. As indicated by the black rectangle in the second row of Fig. \ref{fig:implement_details}, three segment pieces are extracted from the data of a single car, and it is difficult to recognize the car given the data of only one piece. However, by placing each piece of the segment into the background of its surrounding data, the car is easily recognized.
	
	Inspired by the above idea, given a segment $s$, this work generates a sample $x$ containing $s$ as the foreground and its neighborhood data as the background. As illustrated in Fig. \ref{fig:implement_sample}(b), a cuboid centered at $s$ with a size of 2.4 m x 5 m x 5 m (height x width x length) is drawn. The LiDAR points inside the cuboid are extracted, and their pixels on the range image are projected onto a canvas with a size of 256x256 to obtain sample $x$, where each pixel is composed of three channels:
	1) Height: distance to the ground surface mapped to [0,255]. Distances in [0 m,6 m] are mapped linearly to [0,255]. Distance less than 0 m or greater than 6 m are mapped to 0 and 255 respectively; 
	2) Range: distance to the sensor, normalized to [0,255]. 
	3) Intensity: reflectance of the LiDAR point in [0,255].
	
	As a small or distant segment may provide insufficient information for a reliable classification, the following criteria are applied.
	\begin{equation}
	\label{eq:sample_criteria}
	\frac{s_n}{s_d}  >\rho \cap s_n >\sigma
	\end{equation}
	where $s_n$ is the number of LiDAR points of segment $s$, $s_d$ is the distance from the LiDAR sensor to the center point of $s$, and $\rho$ and $\sigma$ are empirically assigned thresholds. Segment $s$ is valid for sample generation if it has more points than $\sigma$ and the n-d ratio is larger than $\rho$. In this research, $\sigma=8.0$ and $\rho=30$.
	
	\begin{table*}
				\centering
		\begin{threeparttable}
		\renewcommand\arraystretch{1.5}			
		\small
		\caption{The data set.}
		\label{tab:dataset}
		
		\begin{tabular}{|p{0.6cm}<{\centering}|p{1.4cm}<{\centering}|c|c|p{0.8cm}<{\centering}|c|c|c|c|c|c|c|c|}
			\hline
			\diagbox[width=3em,height=1.3em]  & Frame  & Dist.(m) & \diagbox[width=3.5em,height=1.3em] & People        & Car            & Cyclist       & Trunk         & Bush           & Building       & Unknown       & Total          \\ \hline
			\multirow{2}{*}{A}     & \multirow{2}{*}{0$\sim$414}     & \multirow{2}{*}{0$\sim$184}   & cons. & 682           & 1897           & 196          & 975           & 3268           & 2450           & 0             & 9468           \\ \cline{4-12} 
			&                                 &                               & anno. & 735           & 2079           & 228          & 1048          & 4330           & 2825           & 2423          & 13668          \\ \hline
			\multirow{2}{*}{B}     & \multirow{2}{*}{414$\sim$829}   & \multirow{2}{*}{184$\sim$350} & cons. & 681           & 1896           & 195          & 977           & 3268           & 2450           & 0             & 9467           \\ \cline{4-12} 
			&                                 &                               & anno. & 736           & 2080           & 227          & 1048          & 4330           & 2825           & 2423          & 13669          \\ \hline
			\multirow{2}{*}{C}     & \multirow{2}{*}{829$\sim$1373}  & \multirow{2}{*}{350$\sim$630} & cons. & 909           & 2529           & 271          & 1301          & 4357           & 3267           & 0             & 14624          \\ \cline{4-12} 
			&                                 &                               & anno. & 980           & 2772           & 304          & 1398          & 5773           & 3767           & 3230          & 18224          \\ \hline
			\multirow{2}{*}{D}     & \multirow{2}{*}{1373$\sim$1838} & \multirow{2}{*}{630$\sim$890} & cons. & 925           & 3542           & 142          & 62            & 2440           & 1254           & 0             & 8365           \\ \cline{4-12} 
			&                                 &                               & anno. & 962           & 4157           & 169          & 89            & 3773           & 1549           & 1667          & 12366          \\ \hline
			\multirow{2}{*}{Total} & \multirow{2}{*}{1838}           & \multirow{2}{*}{890}          & cons. & \textbf{3197} & \textbf{9864}  & \textbf{804} & \textbf{3315} & \textbf{13333} & \textbf{9421}  & \textbf{0}    & \textbf{39934} \\ \cline{4-12} 
			&                                 &                               & anno. & \textbf{3413} & \textbf{11088} & \textbf{928} & \textbf{3583} & \textbf{18206} & \textbf{10966} & \textbf{9743} & \textbf{57927} \\ \hline
			\end{tabular}
		
		\begin{tablenotes}
			 \footnotesize
			\item[1] The A-D corresponding the route in Fig.\ref{fig:datacollect}.
			\item[2] Dist.: the travel distance. cons.: constraint. anno.: annotation.
		\end{tablenotes}
		
	\end{threeparttable}
	\end{table*}

	\begin{figure}
		\centering
		\includegraphics[width=0.49\textwidth]{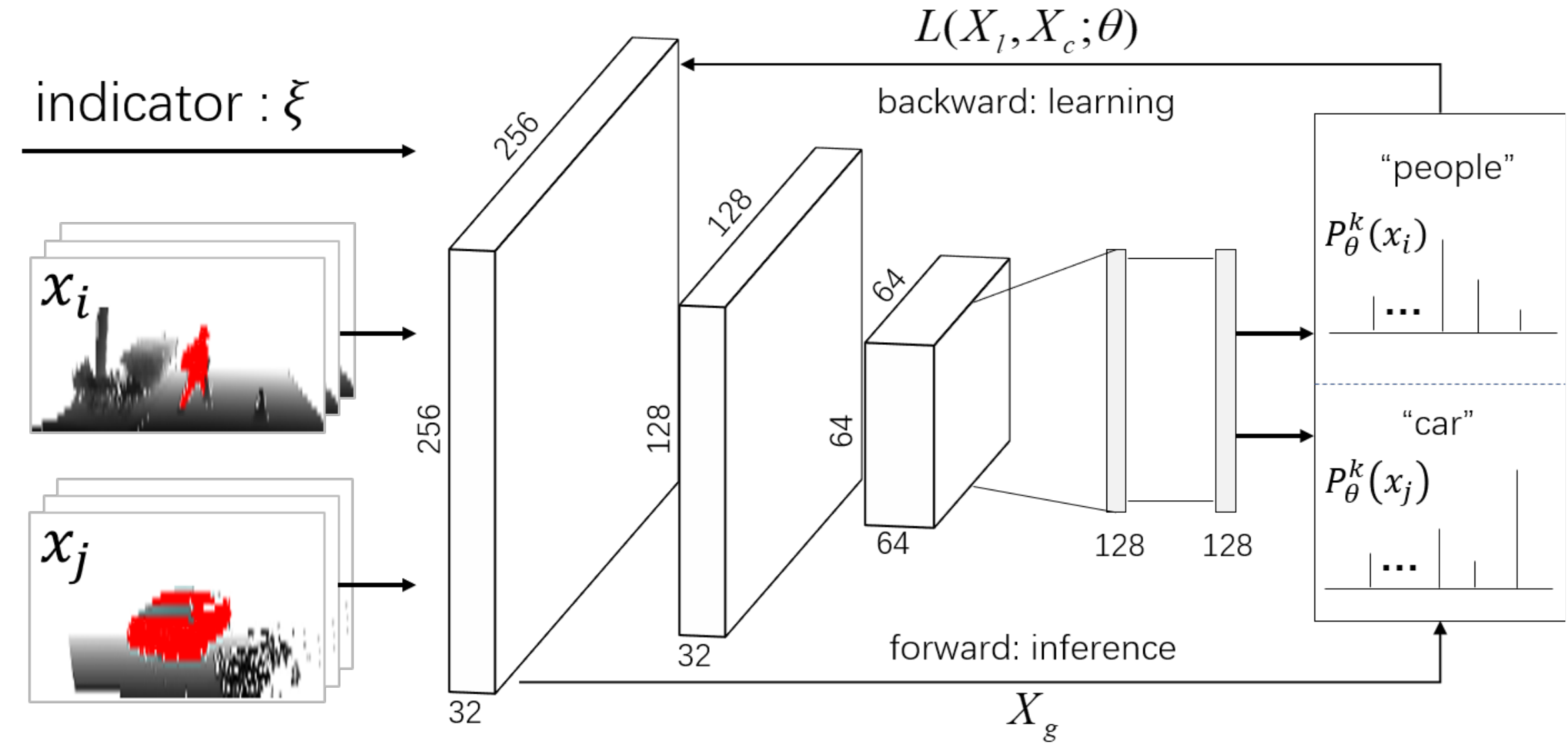}
		\caption{The classifier for offline semi-supervised learning.}
		\label{fig:network}
	\end{figure}
	
	\subsection{Semi-supervised Learning}
	A CNN is used as the classifier $f_\theta$, as shown in Fig. \ref{fig:network}, with specifically designed input and loss function. We rewrite the loss function of Equation (\ref{eq:loss}) as below. 
			\begin{equation}
			\label{eq:loss1}
			\begin{split}
			&L(X_l,X_c;\theta)=\\
			&\left\{
			\begin{array}{lr}
			L_c(X_c;\theta), & M=0,N\neq 0\\
			L_l(X_l;\theta), & M\neq 0,N=0\\
			L_l(X_l;\theta)+L_c(X_c;\theta), & M\neq 0 ,N\neq 0.
			\end{array}
			\right.
			\end{split}
			\end{equation}

	If no supervised samples exists, i.e., $M=0,N\neq 0$, the loss function degenerates to $L_c(X_c;\theta)$, representing unsupervised learning. If there are no constraints, i.e., $M\neq 0,N=0$, the loss function becomes $L_l(X_l;\theta)$, representing supervised learning. Finally, if both supervised samples and constraints exist, i.e., $M\neq 0 ,N\neq 0$, the loss function is in its full form, i.e., $L_l(X_l;\theta)+L_c(X_c;\theta)$, representing semi-supervised learning. The classifier is designed to adapt to all the above cases.
		
	\begin{algorithm}
		\caption{\small the training of the CNN classifier}
		\label{training_classifier}
		\begin{algorithmic}[1]	
			\Require $X_c, X_l$
			\Ensure the classifier parameter $\theta$
			\State Initialize $\varPhi_l, \varPhi_c$ with empty.
			\State Make input pairs:
			\State	\hskip\algorithmicindent $\varPhi_c \leftarrow \langle x_i,x_j;\xi=1\rangle, \forall \langle x_i,x_j \rangle \in X_c$ 
			\State	\hskip\algorithmicindent $\varPhi_l \leftarrow \langle x_i,y_i,x_j,y_j;\xi=0\rangle, \forall \langle x_i,y_i,x_j,y_j \rangle \in X_l$

			\For{\textbf{each} step \textbf{in} \textbf{}training}
				\State$\varPhi^n_c\leftarrow $ take $n$ items from $\varPhi_c, n>0$			
				\State$\varPhi^m_l\leftarrow $ take $m$ items from $\varPhi_l, m>0$
				\State$\varPhi=\varPhi^m_l \cup \varPhi^n_c$
				\For {\textbf{each} item \textbf{in} $\varPhi$}
				\If{$\xi=0$}
					\State $L(X_l,X_c;\theta)=L_l(X_l;\theta)$
					\State	\hskip\algorithmicindent $=L_l(x_i,y_i,x_j,y_j;\theta)$
				\Else
					\State $L(X_l,X_c;\theta)=L_c(X_c;\theta)=L_c(x_i,x_j;\theta)$
				\EndIf
				\State Do backward learning.
				\EndFor
			\EndFor

			\State return $\theta$
			
		\end{algorithmic}
	\end{algorithm}	
	
	To allow both supervised samples and pairwise constraints in model training, the CNN is designed to take two samples $x_i$ and $x_j$ as input and output their labels $y_i$ and $y_j$ simultaneously. An indicator $\xi$ is used to specify whether the two samples are a constrained pair ($\xi=1$) or individuals ($\xi=0$). Hence, the loss functions are converted to the following.
	
	\begin{equation}
		\label{eq:loss3}
		\begin{split}
		L(X_l,&X_c,\xi=1;\theta)=L_l(X_l;\theta)=L_l(x_i,y_i,x_j,y_j;\theta)\\
		&=-\frac{1}{2}\sum_{t=i}^j\sum_{k=1}^K{\textbf{1}[y^{k}_{t}=1]ln(P^k_{\theta}(x_t))};\\
		L(X_l,&X_c,\xi=0;\theta)=L_c(X_c;\theta)=L_c(x_i,x_j;\theta)\\	
		&=\frac{1}{2}(1-\sum_{k=1}^KP^k_{\theta}(x_{i})P^k_{\theta}(x_{j})).	
		\end{split}
	\end{equation}	
	
	In training, the supervised samples $X_l$ and constraints $X_c$ are randomly fed into the CNN, and the model parameters are adjusted in the traditional back-propagation manner. Pseudo code of the training process is given in Algorithm \ref{training_classifier}. As the focus of this research is to learn a classifier using small number of expensive supervised samples and a large number of inexpensive constraints, the unbalanced sample problem should be considered. In this study, we set the number of constraints to 5 times the number of supervised samples, which is described as $n/m=5$ in lines 6-7 of Algorithm \ref{training_classifier}.
	
	In online classification, we have $\xi \equiv 0$, and in the case of only one sample, we have $x_i = x_j$.
	
		\begin{figure}
			\centering
			\includegraphics[width=0.49\textwidth]{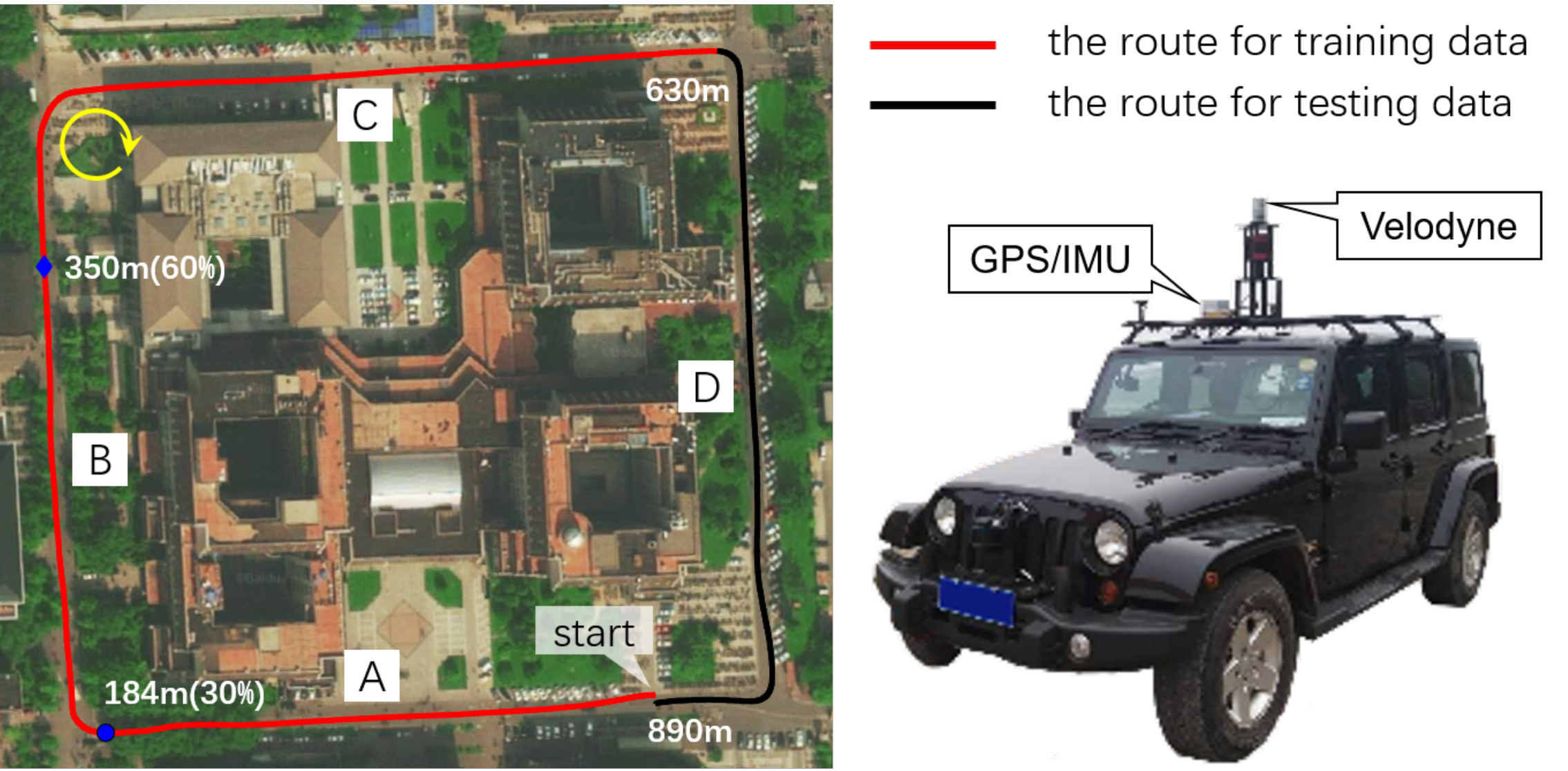}
			\caption{The routes of data collection and the platform configuration. The 30\%/60\% in the left means 30\%/60\% travel of the route for training data.}
			\label{fig:datacollect}
		\end{figure}

	\begin{table*}
		\renewcommand\arraystretch{1.5}
		\centering
		\small
		\caption{The experimental settings on training data.}
		\label{tab:trainingdatasetting}
		\begin{tabular}{|c|c|ccccccc|}
			\hline
			\multicolumn{2}{|c|}{Settings}              & \multicolumn{1}{c|}{People} & \multicolumn{1}{c|}{Car} & \multicolumn{1}{c|}{Cyclist} & \multicolumn{1}{c|}{trunk} & \multicolumn{1}{c|}{Bush} & \multicolumn{1}{c|}{Building} & Unknown \\ \hline
			\multirow{2}{*}{baseline\_min} & constraint & 0                           & 0                        & 0                            & 0                          & 0                         & 0                             & 0       \\
			& annotation & 350                         & 350                      & 350                          & 350                        & 350                       & 350                           & 5650    \\ \hline
			\multirow{2}{*}{baseline\_max} & constraint & 0                           & 0                        & 0                            & 0                          & 0                         & 0                             & 0       \\
			& annotation & 1715                        & 4851                     & 531                          & 2445                       & 10103                     & 6591                          & 5650    \\ \hline
			\multirow{2}{*}{constriant30}  & constraint & 682                         & 1897                     & 196                          & 976                        & 3268                      & 2450                          & 0       \\
			& annotation & 350                         & 350                      & 350                          & 350                        & 350                       & 350                           & 5650    \\ \hline
			\multirow{2}{*}{constriant60}  & constraint & 1363                        & 3793                     & 391                          & 1952                       & 6536                      & 4900                          & 0       \\
			& annotation & 350                         & 350                      & 350                          & 350                        & 350                       & 350                           & 5650    \\ \hline
			\multirow{2}{*}{constriant100} & constraint & 2272                        & 6322                     & 652                          & 3253                       & 10893                     & 8167                          & 0       \\
			& annotation & 350                         & 350                      & 350                          & 350                        & 350                       & 350                           & 5650    \\ \hline
		\end{tabular}
	\end{table*}
	
	\begin{figure*}
		\centering
		\includegraphics[width=0.8\textwidth]{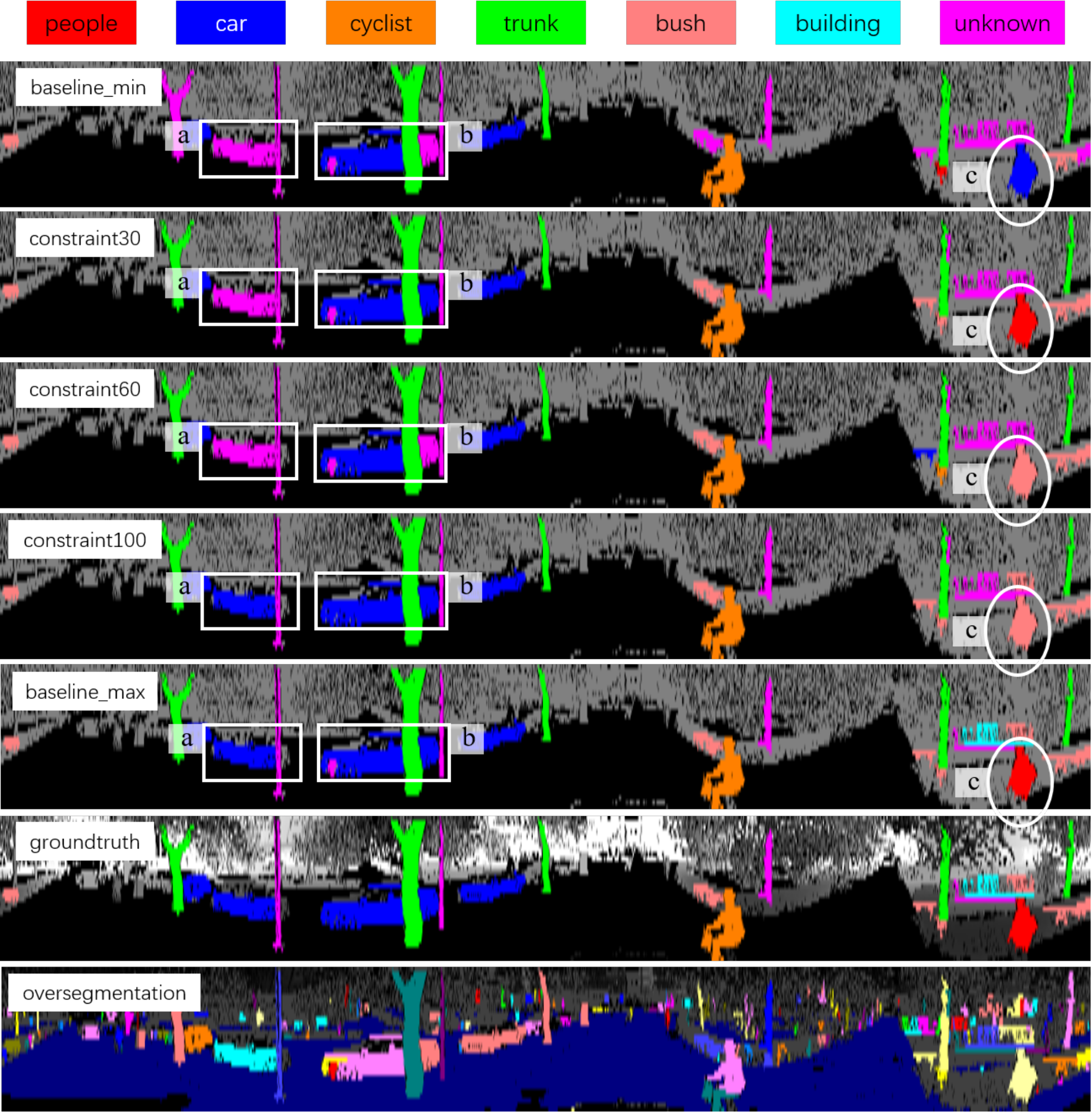}
		\caption{The qualitative results on training data. The rectangles a and b show that the constraint100 has better performance on car, and the ellipse c shows that the constraint100 makes an error classification when the people near the bush.}
		\label{fig:quality_result_training}
	\end{figure*}	
	\begin{table*}[]
		\centering
		\small
		\caption{F measure on training data.}
		\label{tab:result_training}
		\renewcommand\arraystretch{1.5}
		\begin{tabular}{|c|c|ccccccc|}
			\hline
			Learning Method& Classifier       & \multicolumn{1}{c|}{People} & \multicolumn{1}{c|}{Car} & \multicolumn{1}{c|}{Cyclist} & \multicolumn{1}{c|}{Trunk} & \multicolumn{1}{c|}{Bush} & \multicolumn{1}{c|}{Building} & Unknown \\ \hline
			sup. & baseline\_min & 77.2                        & 68.5                     & 66.1                         & 81.0                       & 56.2                      & 78.7                          & 33.9    \\ \hline
			semi-sup. & constraint30  & 77.2                        & 83.0                     & 69.6                         & 86.6                       & 61.9                      & 81.8                          & 36.8    \\ \hline
			semi-sup. & constraint60  & 81.5                        & 85.8                     & 71.3                         & 90.4                       & 80.7                      & 86.6                          & 48.3    \\ \hline
			semi-sup. & constraint100 & \textbf{87.0}                        & \textbf{90.4}                     & \textbf{77.8}                         & \textbf{90.6}                       & \textbf{81.2}                      & \textbf{87.7}                          & \textbf{52.2}    \\ \hline
			sup. & baseline\_max & \textbf{\color{red}93.3}                        & \textbf{\color{red}96.4}                     & \textbf{\color{red}83.8}                      & \textbf{\color{red}96.3}                       & \textbf{\color{red}92.3 }                     & \textbf{\color{red}94.3}                          & \textbf{\color{red}80.0}    \\ \hline
		\end{tabular}
	\end{table*}	

	\section{EXPERIMENTAL RESULTS}
	\subsection{Data Set}
			
	The performance of the proposed method is evaluated on a dynamic campus dataset collected by an instrumented vehicle with a GPS/IMU suite and a Velodyne-HDL32, as shown in Fig. \ref{fig:datacollect}. The total route is approximately 890 meters. All sensor data are collected, and each data frame is associated with a time log for synchronization. The GPS/IMU data are logged at 100 Hz. The LiDAR data are recorded at 10 Hz and include 1373 frames of training data (red line in Fig. \ref{fig:datacollect}) and 465 frames of testing data (black line in Fig. \ref{fig:datacollect}). One frame can produce multiple samples, for example, we obtain 6931 car samples from the 1373 frames of training data in TABLE \ref{tab:dataset}. In total, we obtain 1838 frames of LiDAR data, 39934 constraints and 57927 manual annotations.

	The details of the dataset are listed in TABLE \ref{tab:dataset}. Six labels are used, i.e., person, car, cyclist, trunk, bush and building. These categories are important for driving applications on a campus; other labels, such as pole and cone, are marked as unknown. We do not generate constraints for the unknown label.

	\subsection{Result - Classifier Training}
	\subsubsection{Experimental settings}
	There are five experimental settings for the training data, as shown in Table \ref{tab:trainingdatasetting}. The training data contain 3 parts, i.e., A, B, C in TABLE \ref{tab:dataset}. A total of 70\% of the data are randomly selected for training, and the remaining 30\% are selected for validation. Except baseline\_max, the settings use only a small amount of annotations. Baseline\_min uses 350 annotations without the constraint, while baseline\_max uses all the annotations. Thus, the baseline method works in a supervised manner. In general, baseline\_min sets the low performance bound and baseline\_max sets the high performance bound. Constraint30 uses the same annotations as the baseline\_min but with additional constraints; tail 30 means all constraints are used during 30\% of the travel, i.e., route A from the start to the 184 m point (30\%) in Fig. \ref{fig:datacollect}. constraint60 and constraint100 have similar configurations to constraint30, except for the amount of constraint data.
				
	According to the settings, the five classifiers are learned separately offline. All the classifiers have the same network architecture, as detailed in Set. IV.C. The difference lies in the loss function, where baseline\_min and baseline\_max use only $L_l(X_l;\theta)$ in Equation (\ref{eq:loss}), and the other settings use both $L_l(X_l;\theta)$ and $L_c(X_c;\theta)$. For the offline learning, the classifier's parameters are saved at fixed training steps; then, the parameters that make the loss less than 1e-4 and achieve the best performance on the validation set is selected. For online inference, all classifiers work in the same way.
	
	\subsubsection{Qualitative results}
	As illustrated in Fig. \ref{fig:quality_result_training}, as the number of constraints increases, the car in the rectangle is successfully classified by constraint100, even if occlusions occur. Our method still produces errors, for example, when the person walks near the bush, the constraint100 classifier wrongly annotates the person as a bush. The main reason for this error is that a single sample contains both the foreground and background; if the background occupies more information than the foreground, the classifier is likely to assign the background label to the sample. 
	
		\begin{figure}
			\centering
			\includegraphics[width=0.5\textwidth]{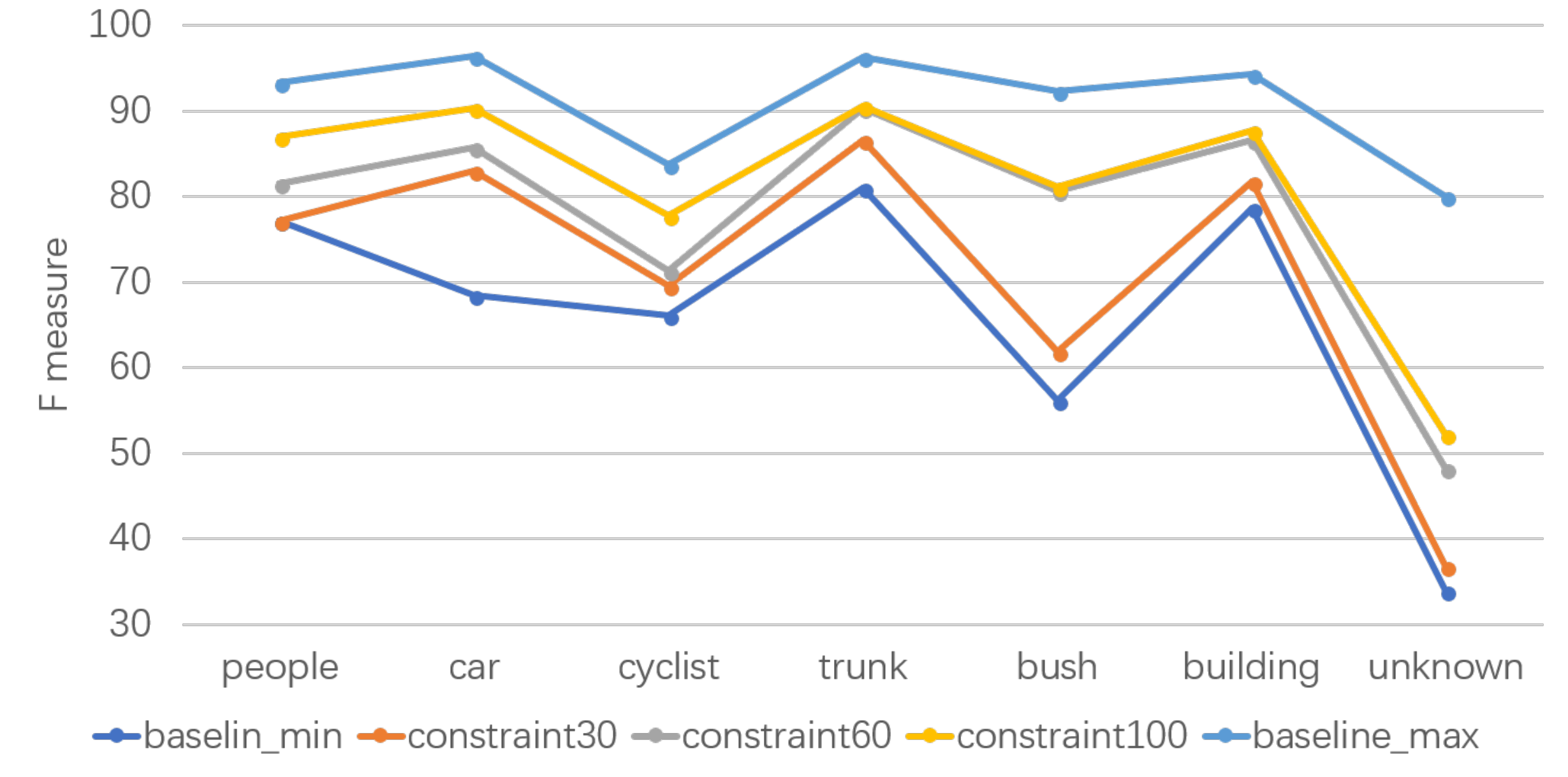}
			\caption{The quantitative comparison on training data.}
			\label{fig:plot_training_result}
		\end{figure}
		
	\subsubsection{Quantitative results}
	The F-measure is adopted for quantitative evaluations and is defined as:
	\begin{equation}
	\label{eq:f-measure}
	F-Measure = \frac{2*recall*precision}{recall+precision}\cdot{100\%}.
	\end{equation} 
	We use the five classifiers to annotate the training set and the validation set, and the quantitative results are shown in TABLE \ref{tab:result_training}. 

	The baseline\_min and baseline\_max results show that a large number of annotations is important for supervised learning. The baseline\_min and constraint100 results indicate that semi-supervised learning is effective: the F-Score of constraint100 increases by 15\% on average. Furthermore, as the number of constraints increases, the performance of the classifier is enhanced. A more intuitive comparison is illustrated in Fig. \ref{fig:plot_training_result}. Although constraint100 is not as good as baseline\_max, in which all annotations are used, it shows promising results, indicating that adding constraints improves the performance of the classifier and reduces the need for annotations. In conclusion, semi-supervised learning, where only a few annotations are used, is effective for 3D point cloud semantic segmentation.
	
	\subsection{Result - Classifier Testing}

	\begin{table*}[]
		\centering
		\small
		\caption{The pre-trained classifiers on testing data.}
		\label{tab:pre-trained-result-testing}
		\renewcommand\arraystretch{1.5}
		\begin{tabular}{|c|c|ccccccc|}
			\hline
			Learning Method & Classifier       & \multicolumn{1}{c|}{People} & \multicolumn{1}{c|}{Car} & \multicolumn{1}{c|}{Cyclist} & \multicolumn{1}{c|}{Trunk} & \multicolumn{1}{c|}{Bush} & \multicolumn{1}{c|}{Building} & Unknown       \\ \hline
			sup. & baseline\_min & 57.6                        & 64.0                     & 26.3                         & 37.5                       & 33.2                      & 49.7                          & 33.0          \\ \hline
			semi-sup. & constraint100 & \textbf{69.4}               & \textbf{81.6}            & \textbf{\color{red}39.8}                & \textbf{\color{red}42.1}              & \textbf{37.3}             & \textbf{55.7}                 & \textbf{39.0} \\ \hline
			sup. & baseline\_max & \textbf{\color{red}73.7}               & \textbf{\color{red}88.8}            & \textbf{30.0}                & \textbf{30.8}              & \textbf{\color{red}71.0}             & \textbf{\color{red}65.4}                 & \textbf{\color{red}53.0} \\ \hline
		\end{tabular}
	\end{table*}
	
	\begin{table*}[]
		\centering
		\small
		\caption{The setting of fine tuning}
		\renewcommand\arraystretch{1.5}		
		\label{tab:fine-tuning-setting}
		\begin{tabular}{|c|ccccccc|}
			\hline
			\diagbox[width=7em,height=1.2em]              & \multicolumn{1}{c|}{People} & \multicolumn{1}{c|}{Car} & \multicolumn{1}{c|}{Cyclist} & \multicolumn{1}{c|}{Trunk} & \multicolumn{1}{c|}{Bush} & \multicolumn{1}{c|}{Building} & Unknown \\ \hline
			anchor sample & 20                          & 20                       & 20                           & 20                         & 20                        & 20                            & 100     \\ \hline
			constraint    & 925                         & 3542                     & 142                          & 62                         & 2440                      & 1254                          & 0       \\ \hline
		\end{tabular}
	\end{table*}

	\begin{table*}[]
		\centering
		\small
		\renewcommand\arraystretch{1.5}
		\caption{F measure on testing data}
		\label{tab:fine-tuning-result}
		\begin{tabular}{|c|c|ccccccc|}
			\hline
			Learning Method & Classifier                   & \multicolumn{1}{c|}{People} & \multicolumn{1}{c|}{Car} & \multicolumn{1}{c|}{Cyclist} & \multicolumn{1}{c|}{Trunk} & \multicolumn{1}{c|}{Bush} & \multicolumn{1}{c|}{Building} & Unknown       \\ \hline
			sup. & baseline\_min             & 57.6                        & 64.0                     & 26.3                         & 37.5                       & 33.2                      & 49.7                          & 33.0          \\ \hline
			semi-sup. & constraint100             & 69.4                        & 81.6                     & 39.8                         & 42.1                       & 37.3                      & 55.7                          & 39.0          \\ \hline
			semi-sup. & constraint100+fine tuning & \textbf{\color{red}77.1}               & \textbf{\color{red}90.7}            & \textbf{\color{red}60.2}                & \textbf{\color{red}55.7}              & \textbf{58.0}             & \textbf{62.3}                 & \textbf{\color{red}54.3} \\ \hline
			sup. & baseline\_max             & \textbf{73.7}               & \textbf{88.8}            & \textbf{30.0}                & \textbf{30.8}              & \textbf{\color{red}71.0}             & \textbf{\color{red}65.4}                 & \textbf{53.0} \\ \hline
		\end{tabular}
	\end{table*}

	In a fixed scene or dataset, a classifier based on supervised learning can easily achieve high performance due to over-fitting. When the classifier is applied to a new scene, additional annotations are necessary to prevent performance degradation. However, large quantities of fine annotations in each new scene are difficult to obtain for driving applications. Thus, how to enhance the adaptability with a few or no new annotations is crucial for practical applications. We detail three experiments in the following to demonstrate the adaptability of the proposed semi-supervised method. The F-measure in Equation (\ref{eq:f-measure}) is used for the quantitative comparison.
	
	\subsubsection{The pretrained result}
	The pretrained classifiers based on training data are directly applied to the testing data, and the results are shown in TABLE \ref{tab:pre-trained-result-testing}. The baseline\_min and constraint100 results show that adding constraints improves the adaptability to the new scene: the F-Score of constraint100 increases by 9\% on average. The constraint100 and baseline\_max results show that constraint100 has higher scores for the cyclist and trunk categories, despite having lower performance in all categories on the training data (TABLE \ref{tab:result_training}). 
	\subsubsection{The unsupervised result}
\begin{figure}
		\centering
		\includegraphics[width=0.3\textwidth]{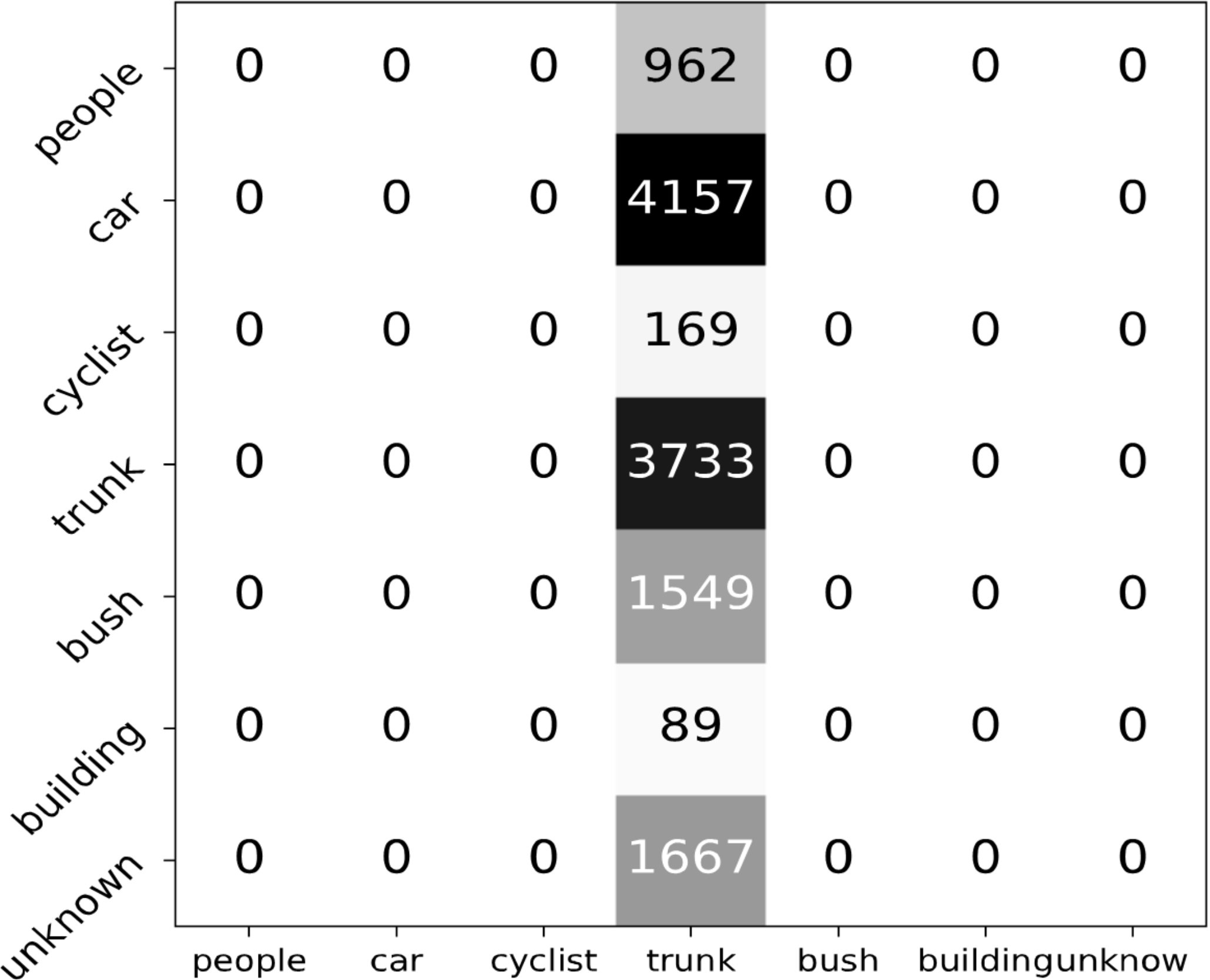}
		\caption{The confusion matrix of unsupervised learning on testing data.}
		\label{fig:cm_unsupervised}
	\end{figure}

	For the new scene, an attempt to use only constraints and no new annotations to improve the pretrained classifier is interesting. Thus, the loss function in Equation. (\ref{eq:loss}) is rewritten as:
	\begin{equation}
		\label{eq:new_loss}
		\begin{split}
		L(X_l,X_c;\theta)&=L_c(X_c;\theta)= \\
		&\frac{1}{N}\sum_{i=1}^N(1-\sum_{k=1}^KP^k_{\theta}(x_{i})P^k_{\theta}(x_{j})).
		\end{split}
	\end{equation}
	Here, the pretrained constraint100 is treated as the initial classifier, and only constraints are used to retrain the model. The results of this unsupervised learning procedure are shown in Fig. \ref{fig:cm_unsupervised}. Regardless of the samples, the classifier erroneously assigns them as a trunk. We can explain this situation with the loss function in Equation (\ref{eq:new_loss}): the constraint loss only penalizes the classifier when it gives $x_{i}$ and $x_{j}$ different labels, so unsupervised learning fails.		

	\begin{figure}
		\centering
		\includegraphics[width=0.5\textwidth]{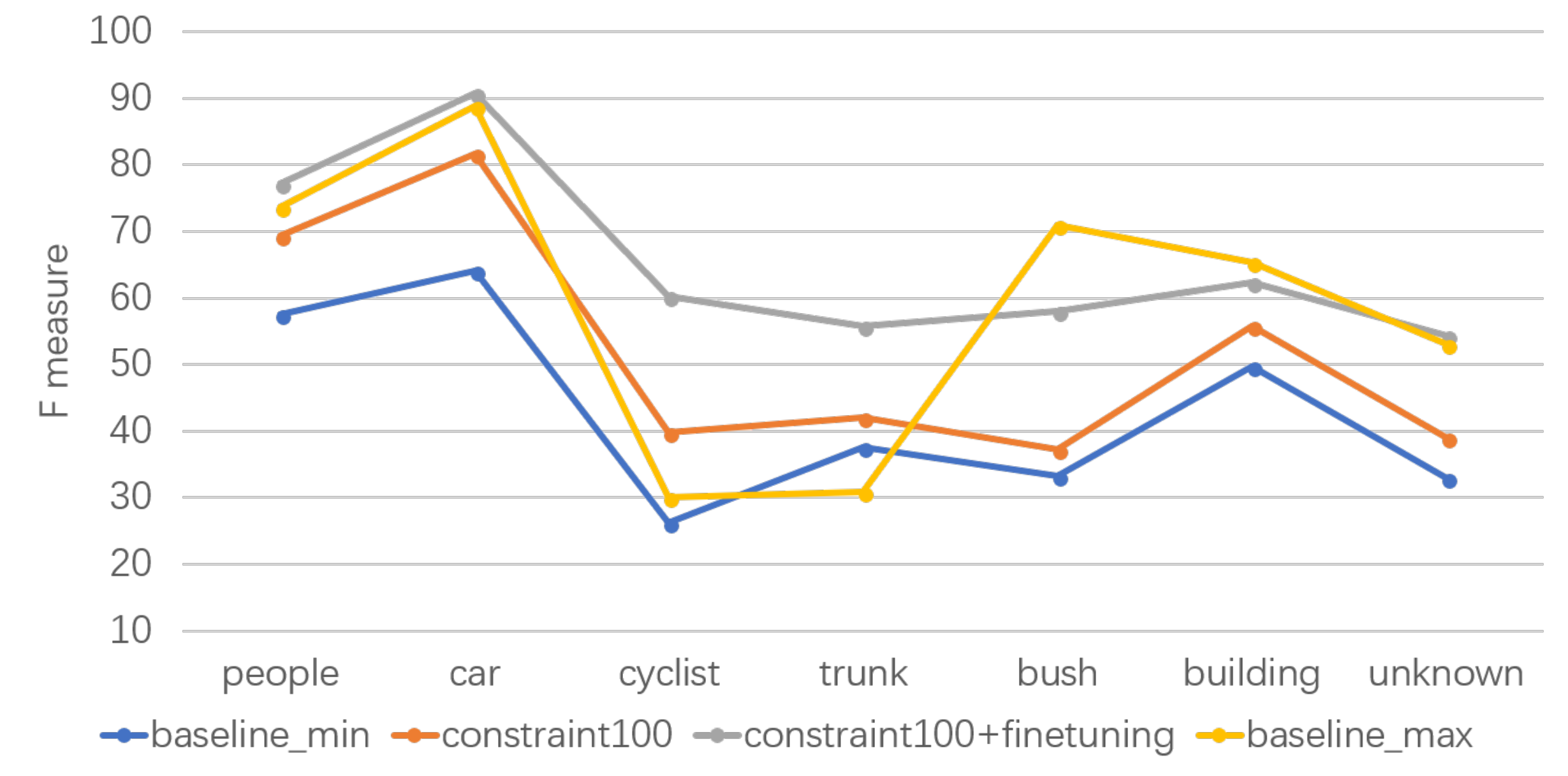}
		\caption{The quantitative comparison on testing data.}
		\label{fig:plot_testing_result}
	\end{figure}		

	\subsubsection{The fine tuning result}
	After finding that unsupervised learning does not work for our method, we attempt to add a few new annotations. Specifically, the new annotations are generated in an interactive way. The pretrained constraint100 is used as the initial classifier to produce classification results on the testing data. Although the pretrained results show low F-Scores in TABLE \ref{tab:pre-trained-result-testing}, a few new annotations can be selected by human confirmation. In this way, we obtain 20 annotations for each category and 100 for the unknown label, as shown in TABLE \ref{tab:fine-tuning-setting}, where the new annotation is renamed the anchor sample.
	
	\begin{figure*}
		\centering
		\includegraphics[width=0.8\textwidth]{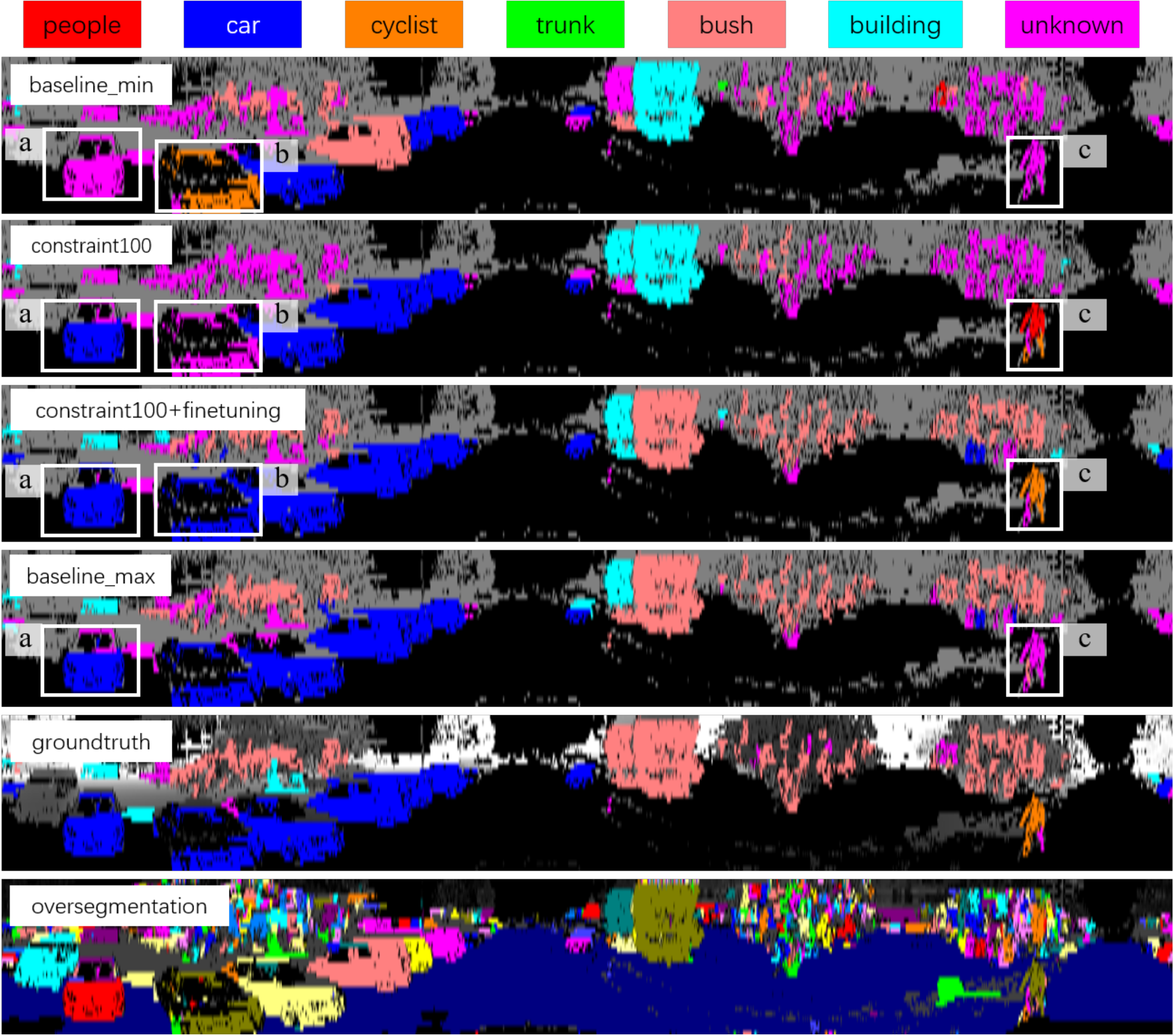}
		\caption{The qualitative results on testing data. The a and b show the comparison on car and c shows that the cyclist is successfully classified after fine-tuning.}
		\label{fig:quality_result_testing}
	\end{figure*}
		
	The pretrained constraint100 model is fine-tuned by combining the anchor sample and constraint. The final quantitative results are shown in TABLE \ref{tab:fine-tuning-result} and Fig. \ref{fig:plot_testing_result}. Comparing constraint100 and the fine-tuned version, the F-Score of the latter increases by 13\% on average, which shows that fine-tuning is an effective way to improve adaptability, even in a semi-supervised manner. Comparing the baseline\_max and the fine-tuned version, the latter has higher scores except on bush and building. The qualitative results are shown in Fig. \ref{fig:quality_result_testing}. In conclusion, fine-tuning with the anchor sample increases the adaptability of the classifier to a new scene.
	
	\section{conclusion and future work}
	A semantic segmentation method for 3D point clouds (i.e., from LiDAR senors) is developed in this research, and semi-supervised learning is utilized to reduce the considerable requirement for fine annotations. The pairwise constraints between adjacent frames are generated via inter-frame data association, and a loss function is designed to help the constraint data to obtain the same label. This method is examined extensively on a new dataset. The superior results indicate improvements in effectiveness and adaptability. Future work will address how to define the sample because including both foreground and background information in the sample can confuse the classifier. In addition, the introduction of new constraints will also be studied.
	\bibliographystyle{IEEEtran}
	\bibliography{myref}
	
\end{document}